\documentclass[letterpaper, 10 pt, conference]{ieeeconf}  

\IEEEoverridecommandlockouts                              

\usepackage{cite}
\usepackage{amsmath} 
\usepackage{amssymb}  
\usepackage{bbm}
\usepackage{bm}
\usepackage{amsmath}
\usepackage{algorithm}
\usepackage{algpseudocode}
\usepackage{graphicx}
\usepackage{multirow}
\usepackage{array}
\usepackage{xcolor}
\usepackage{dsfont}
\usepackage{booktabs}
\usepackage[justification=justified]{caption}

\title{\LARGE \bf
Language-guided Viewpoint and Object Manipulation Planning for Active Sensing in Confined Environments
}

\title{\LARGE \bf
Language-guided Active Sensing via Object Manipulation in Confined, Cluttered Environments
}

\title{\LARGE \bf
Language-guided Active Sensing of Confined, Cluttered Environments via Object Rearrangement Planning
}

\author{Weihan Chen, Hanwen Ren, and Ahmed H. Qureshi 
\thanks{*This work was supported by the National Science Foundation (NSF) under award no. 2204528.}
\thanks{Weihan Chen, Hanwen Ren, and Ahmed H. Qureshi are with the Department of Computer Science, Purdue University, West Lafayette, IN, USA, 47907. Email {\tt\small$\{$chen3189, ren221, ahqureshi$\}@$purdue.edu}}%
}

\begin{document}

\newcommand{\R}[0]{\mathbb{R}}
\newcommand{\langdir}[0]{\mathcal{D}}
\newcommand{\objcoord}[0]{\boldsymbol{x}}

\maketitle
\thispagestyle{empty}
\pagestyle{empty}

\begin{abstract}
Language-guided active sensing is a robotics subtask where a robot with an onboard sensor interacts efficiently with the environment via object manipulation to maximize perceptual information, following given language instructions. These tasks appear in various practical robotics applications, such as household service, search and rescue, and environment monitoring. Despite many applications, the existing works do not account for language instructions and have mainly focused on surface sensing, i.e., perceiving the environment from the outside without rearranging it for dense sensing. Therefore, in this paper, we introduce the first language-guided active sensing approach that allows users to observe specific parts of the environment via object manipulation. Our method spatially associates the environment with language instructions, determines the best camera viewpoints for perception, and then iteratively selects and relocates the best view-blocking objects to provide the dense perception of the region of interest. We evaluate our method against different baseline algorithms in simulation and also demonstrate it in real-world confined cabinet-like settings with multiple unknown objects. Our results show that the proposed method exhibits better performance across different metrics and successfully generalizes to real-world complex scenarios.
\end{abstract}

\section{Introduction}
Language-guided active sensing is an integrated perception, planning, and control problem wherein a robot, equipped with onboard perception sensors, optimally collects the perceptual information based on given language commands by interacting with the surrounding environment \cite{1250604}. In confined and narrow-passage environments, these tasks often require adept object manipulation to effectively collect visual data and provide the desired dense perception. These scenarios appear in various settings such as in-home assistance \cite{8665127, Umari2017AutonomousRE} where the disabled and elderly people will rely on robot systems to provide them with the perception of confined environments such as cabinets, drawers, or refrigerators. Likewise, in search and rescue \cite{8606991, sar2}, a remote human operator may instruct a robot to provide visuals of specific areas by manipulating the underlying objects. Despite crucial and essential applications of language-guided active sensing, such tasks remain in an unexplored area of research. Most existing work focuses on active sensing of open, table-top-like environments, with an exception of \cite{10101696}, which considers confined environments. Still, none of these methods rearrange objects for dense sensing or account for language instructions. 



Therefore, this paper presents a language-guided active sensing system that can effectively provide a dense perception of the given unknown, cluttered environments by manipulating objects to meet a specified requirement of a user given by natural language commands. Our method aims to meet the requirements of a user prompt by exploring an unknown environment with minimal camera movements and object manipulations. The main contributions of our framework are summarized as follows:

\begin{itemize}
\item Language to environment spatial correspondence matching module to identify the user's region of interest (ROI) for active perception in the given environment.
\item An MPC-style viewpoint planning method that uses Gaussian Mixture Models (GMM) with a novel neural network-based utility function to select the best viewpoints for observing the ROIs.
\item An analytical approach to identify ROI view-blocking objects for relocation via manipulation to clear the way for selected viewpoints to observe the user's ROI.
\item A unified language-guided active sensing system that combines the above methods for fast selection of viewpoints and relocation of view-blocking objects to complete the user command of sensing a specific portion of the environment. 
\end{itemize}
\section{Related work}
The most relevant works to our approach are on the topic of autonomous exploration and next-best view selection. The traditional and prominent approaches to autonomous exploration include frontier-based and search-based methods. The frontier-based autonomous exploration method is first introduced in \cite{dudek2010computational}, where the robot explores the boundary between mapped and unmapped space. Later work expanded on this technique to achieve a higher coverage along the path of the frontier \cite{heng2015efficient, gonzalez2002navigation, cieslewski2017rapid}. However, these frontier-based methods require overlapping of sequential observations. This would lead to too many viewpoints being generated and not suitable for efficient completion of user's requests. In contrast, the search-based method utilizes a utility function to guide the search, first introduced in \cite{Connolly1985TheDO}. This method is most prominent in 3D reconstruction settings. Further methods derived from this include \cite{VasquezGomez2014ViewPF, VasquezGomez2017ViewstatePF, Isler2016AnIG}, which propose different utility measures that can better maximize the coverage. These methods are generally used in single-object reconstruction settings with assumptions about the underlying geometry, which is unfavorable for narrow and cluttered settings.

More recent data-driven approaches leverage deep neural networks to approximate the utility function for the next best view generation \cite{MENDOZA2020224, hepp2018learntoscore, Gallos2019ActiveVI}. The introduction of neural networks into next-best view estimation eliminates the need to physically move the camera, which reduces sub-optimal movements. These methods, however, only operate on single objects or open spaces, which is unsuitable for exploration in a multi-object, narrow environment. The most recent approach that explores confined, narrow environments is VPFormer \cite{10101696}, which produces results that show minimal viewpoint generation for scene coverage. However, this approach assumes a scene that is observable solely by moving the camera, i.e., without object manipulation. 

In contrast to the above methods, we aim for settings where objects can prevent observation of significant parts of the scene, making selection and relocation of view-blocking objects necessary. Furthermore, we also consider human-in-the-loop settings where they provide language commands to indicate the region of interest in the environment for dense active sensing. Although natural language as commands have been used in various robotic tasks \cite{Kollar2010TowardUN, ahn2022i, Tellex2011UnderstandingNL}, their role in performing active sensing is unexplored. Similarly, recent studies enable the manipulation of unknown objects using point cloud data gathered from stereo or RGB-D cameras \cite{5650206, mittal2022articulated, kollar2021simnet} but not from the perceptive of rearranging scenes for better visual coverage. To the best of our knowledge, this work presents the first framework for language-guided active sensing via object manipulation in confined, narrow environments.


\begin{figure*}[t]
\includegraphics[width=\textwidth]{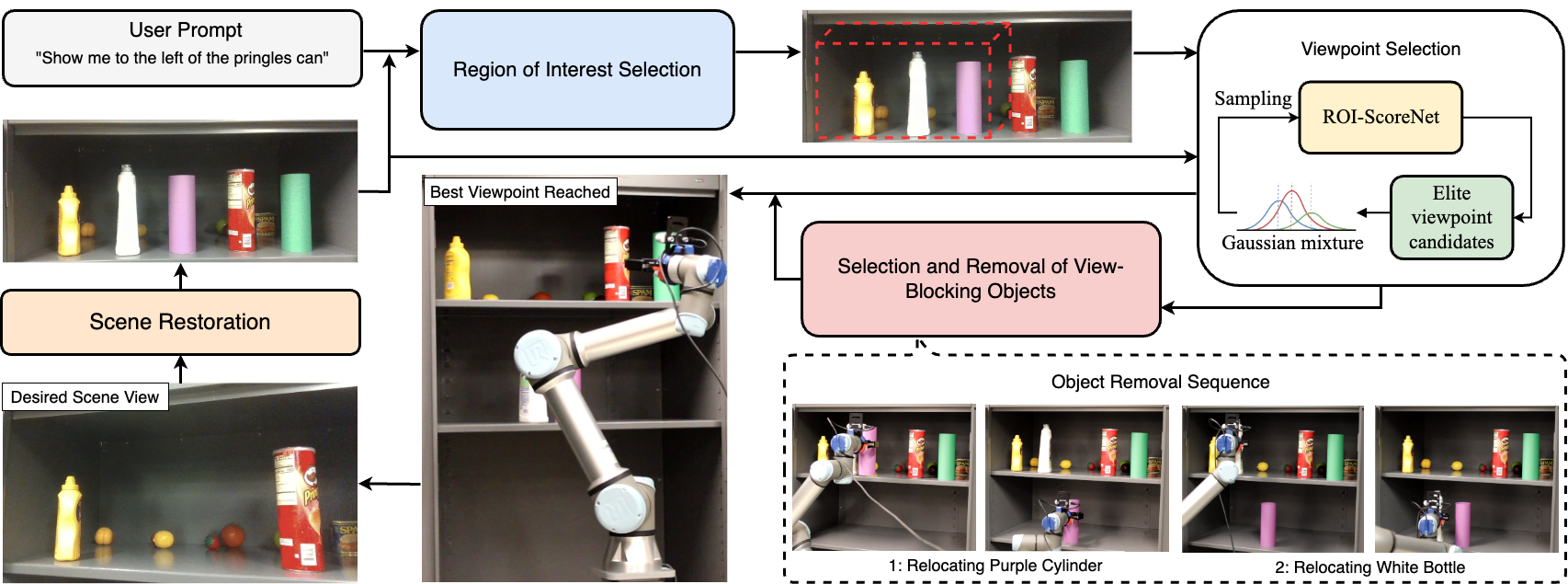}
\centering
\caption{This figure shows the language-guided active sensing pipeline with an example user prompt. The user prompt and initial scene are processed by the language linking module for the region of interest selection. With the region of interest, the MPC-GMM viewpoint selection algorithm uses ScoreNet and Gaussian mixture to select the optimal viewpoint for the environment setup. After moving to the optimal viewpoint, the view-blocking objects are removed sequentially using the blocking score formulation. Finally, the modified scene view is shown to the user. The system will move on to the next viewpoint if the desired coverage is not reached.}
\label{fig1:pipeline}
\vspace{-0.2in}
\end{figure*}

\section{Proposed Methods}
This section formally presents the main components of our language-guided active sensing via object manipulation system with the overall pipeline shown in Fig. \ref{fig1:pipeline}.  

\subsection{Problem Formulation}
Let a confined environment be represented as $S \subseteq \mathbb{R}^3$, whose ranges along the $x, y,$ and $z$ axes are assumed to be known and denoted as $d_x, d_y,$ and $d_z$, respectively. The value of a coordinate $S(i, j, k)$ is either 1 or 0, where the collection of the former forms the observed region $S_o$ and that of the latter forms the unobserved region $S_{ou}$. The scene also has a collection of household items $\{o_1, o_2,...\}$. A robot arm equipped with an in-hand RGB-D camera placed in front of the scene takes an initial image $I^0$ 
and presents it to a human user. The user then describes, via language command, a region of interest (ROI), $S_{ROI} \subseteq S$, that requires active visual exploration. In order to measure how well the user's request is fulfilled, a coverage rate of the observed region $S_o$ over the user-specified region $S_{ROI}$ is defined as:
\begin{equation}
    \phi(S_o, S_{ROI}) = \frac{\sum_{i}^{dx}\sum_{j}^{dy}\sum_{k}^{dz}(\mathds{1}_{S_{ROI} \cap S_o}(i,j,k))}{\sum_{i}^{dx}\sum_{j}^{dy}\sum_{k}^{dz}\mathds{1}_{S_{ROI}}(i,j,k)}
\end{equation}
The region of interest $S_{ROI}$ is fully explored when $\phi(S_o, S_{ROI})\approx 1$. At any time step $t$, the robot can remove a detected view-blocking object $o^t$ from the scene and navigate the camera to a viewpoint $v^t \in \mathcal{V} \subseteq SE(3)$. Each camera viewpoint $v^t$ after an object $o^t$ relocation leads to an image $I^{t+1}$ and a scene observation $S^{t+1}$ with newly obtained information marked as $S^{t+1}_{o}(o^t, v^t)\backslash S^{t}_o$. In all, the objective of the problem is to maximize the overall scene coverage rate $\phi(S_o, S_{ROI})$ leveraging two policy functions $\pi_o$, and $\pi_v$, that makes the best choice of the object manipulation and viewpoint selection at any given step, resulting in the shortest robot action sequence. The objective can be formally written using the min-max formulation:\vspace{-0.1in}
\begin{equation}
    \max_{v^t\sim\pi_v, o^t\sim\pi_o} \min_{T} \sum_{t=0}^{T-1} \phi(S_o^{t+1}(o^t, v^t)\backslash S_o^t, S_{ROI})
\vspace{-0.05in}\end{equation}
During the active ROI sensing procedure, all images $I^t$ are sent to the user's end as a response. If the robot agent achieves the objective, it also means that the user can see all the details in the ROI using the minimal number of images, which fulfills the request.\par
The remainder of the method section discusses our design to optimize the above objective function. Furthermore, to describe various functions of our framework, we use a notation $A_{\{B\}}$ denoting a list $A$ composed of $B$ elements for brevity.
  
\subsection{Language-guided ROI Generation}
The user makes a language prompt to describe a region relative to an observed object for active sensing, e.g., ``Show me to the left of the pink cylinder." Our language linker then identifies the anchor object $o_i$ and a relative spatial direction $l$, as shown by the red-dotted box in Fig. \ref{fig1:pipeline}. More specifically, we use \cite{spacy} to parse the user query into part of speech tags, and then $o_i$ is obtained by extracting the tokens with a tag of \textit{pobj} (object of preposition) or \textit{dobj} (direct object). On the other hand, the direction is identified by iterating through the sentence and finding tokens whose synonyms include a directional word. The extracted direction $l$ resides in the set containing $\{\text{left, right, behind, front}\}$, corresponding to both sides of the $X$ and $Y$ axes. Let the location of the anchor object $o_i$ be represented as $r(o_i) = (x_{o_i},y_{o_i},z_{o_i})$, the region of interest $S_{ROI}$ is constructed by expanding $r_{o_i}$ in all axes until the boundaries of the environment following the user-specified spatial direction $l$.


\subsection{ROI-ScoreNet}
Due to the irregular shape and various potential locations of the multiple objects in the scene, estimating the new information gained from a given camera pose is nontrivial. Thus, we propose an ROI coverage score function, called ROI-ScoreNet, that predicts the possible scene coverage gain from a given camera viewpoint candidate without physcially repositioning the camera. This function is inspired from the ScoreNet in \cite{10101696}.
However, the original ScoreNet formulation considers the entire scene instead of a ROI. Thus, we proposed a new neural network structure for ROI-ScoreNet. The ROI-ScoreNet can be viewed as a function $f_\theta$ with parameters $\theta$ that estimates the coverage rate $\hat{c}\in [0, 1]$ over an ROI given the current scene $S^t$, the region of interest $S_{ROI}$, and any arbitrary viewpoint $v^t$.
\begin{equation}
    \hat{c}\leftarrow f_\theta (S^t, S_{ROI}, v^t)
\end{equation}
The ROI-ScoreNet is comprised of a two-channel 3DCNN layer \cite{tran2015learning} that embeds the scene representation $S^t$ and region of interest $S_{ROI}$ into a latent space $Z_S$. The viewpoint $v^t$ is passed into a fully connected neural network to generate the latent viewpoint embedding $Z_v$. These latent features are concatenated and passed into another fully connected neural network to generate the estimated scene coverage $\hat{c}$. The ROI-ScoreNet is trained in a supervised manner using a dataset composed of $(S^t, S_{ROI}, v^t, c^{t+1})$ pairs from scenes with various objects in the simulation environment. The dataset is generated by moving the in-hand camera into a random viewpoint and recording the resulting coverage over the region of interest. During training, the objective is to minimize the Mean Squared Error (MSE) between the predicted coverage rate $\hat{c^{t+1}}$ and the ground truth $c^{t+1}$, i.e., $\frac{1}{M}\sum_{i=1}^{M}||c^{t+1}-\hat{c}^{t+1}||^2$.

\subsection{GMM-MPC Viewpoint Generation}
While the ROI-ScoreNet allows fast estimation of scene coverage rate for various viewpoints around the scene, the remaining challenge is to devise an underlying algorithm that can effectively select the optimal one. Thus, we propose GMM-MPC, a bilevel MPC-style viewpoint generation algorithm derived from \cite{10101696, bharadhwaj2020modelpredictive}. 
Our GMM-MPC extends the bilevel MPC method by replacing the single Gaussian distribution with a Gaussian Mixture Model (GMM). The rationale behind this is the multi-modal nature of the best viewpoint distributions, as there can be many optimal viewpoints for a given ROI. By sampling only around one initial point, the resulting coverage is more likely to be a local maximum. The pseudocode of the GMM-MPC viewpoint generation method is shown in Algorithm \ref{alg1}. In the first stage, GMM-MPC selects $N$ highest scoring viewpoints from uniformly generated samples around the scene as the initial centroids $\mu_1,...,\mu_N$ of the Gaussian mixture model (lines 2-5). The algorithm then keeps sampling from the existing distribution, selecting the $K$ elite samples determined by the ROI-ScoreNet, and refining each of the $N$ Gaussian distributions in the model for $M$ iterations (lines 6-10). Finally, the best viewpoints from the last round are returned as the candidates of the best next viewpoint (line 12). The optimal number of components $N$ for the Gaussian mixture is chosen via cross-validation and set to seven, which yields better performance in general while maintaining a reasonable time consumption.
\begin{algorithm}
    \caption{GMM-MPC viewpoint generation algorithm}
    \begin{algorithmic}[1]
        \State{initialize $\sigma_{\{N\}}$}
        \State{$v^t_{\{S\}}\sim$ \text{Uniform}$(v_s \in S^t_o)_{\{S\}}$}
        \State{$\hat{c}^t_{\{S\}}\leftarrow f_\theta(S^t,S_{ROI},v^t_{\{S\}})$} \Comment{ROI-ScoreNet selection}
            \State{$(v^t, \hat{c})_{\{N\}}\leftarrow \text{Sort}(v^t_{\{S\}}, \hat{c}^t_{\{S\}})[:N]$}
        \State{$\mu_{\{N\}}\leftarrow  v^t_{\{N\}}$} \Comment{Initial GMM centroids}
        \For{$iter\leftarrow 1$ to $M$}
            \State{$v^t_{\{B\}}\sim \text{GMM}(\mu_{\{N\}},\sigma_{\{N\}})$}
            \State{$\hat{c}^t_{\{B\}}\leftarrow f_\theta(S^t,S_{ROI},v^t_{\{B\}})$}
            \State{$(v^t, \hat{c})_{\{K\}}\leftarrow \text{Sort}(v^t_{\{B\}}, \hat{c}^t_{\{B\}})[:K]$} \Comment{Elite samples}
            \State{$\mu_{\{N\}},\sigma_{\{N\}}\leftarrow Fit(v^t_{\{K\}})$}
        \EndFor
        \State{\Return{$v^t_{\{K\}}$}}\Comment{Best viewpoint}
    \end{algorithmic}
    \label{alg1}
\end{algorithm}
\subsection{View-Blocking Object Selection}
In the cluttered environment setting, the ROI cannot be fully covered by simply changing viewpoints but requires object manipulation. In order to determine which object to remove at the current step, we proposed the following view-blocking rate estimation method to guide the process further. For each observed object $o_i$, a voxelized point cloud is extracted using the depth image. Then, for each point $p_i$ in the point cloud, a ray $r$ is cast from the current camera viewpoint $v^{t}$ and shoots towards it. As the ROI is a convex region, the ray will intersect it at two points, denoted $p_{i1}$ and $p_{i2}$, with $p_{i2}$ being the point further away from the camera. A viewpoint rate $h$ associated with every object is calculated as the sum of line segment lengths:\vspace{-0.05in}
\begin{equation}
    h_{o_i}=\sum_{p_i \in o_i}||p_i-p_{i2}||_2\mathds{1}_{(p_{i2}-p_{i})\cdot(v^{t-1}-p_i) < 0}(p_i)
\vspace{-0.1in}\end{equation}
The indicator term excludes points that are outside the region of interest and therefore does not contribute to blocking the view. Such points would have $p_{i2}$ between the camera and $p_i$, causing the stated dot product to be positive. The rationale behind the proposed method is that the view-blocking volume is positively correlated to the integration of the line segment lengths from the objects' surface to the boundaries of the ROI following the viewpoint's orientation. We will use the notation \texttt{BScore($S^t, S_{ROI}, v^t$)} to denote a function that computes the blocking score $h_{o_i}$ for every object $o_i$ that is currently observable. During execution, a max heap is kept during the object manipulation phase. The heap will be initialized with the blocking score of all observed objects with a blocking score greater than a minimum threshold. At any step, the robot arm removes the first object that popped out of the heap, which blocks the most view.  
\subsection{Language-guided Active Sensing Pipeline}
Algorithm \ref{NG-AS algorithm} describes our language-guided active neural sensing via object manipulation approach. The system starts with an initial viewpoint $v^0$ that looks directly at the scene's center. An overview image $I^0$ taken from this initial view is presented to the user to obtain a language prompt (lines 2-4). Once the prompt is received, the ROI-Generation module creates the corresponding region of interest $S_{ROI}$ (Line 5). Next, the pipeline actively explores the ROI through visual perception and object manipulation. During this phase, the best viewpoint $v^t$ is predicted by the GMM-MPC viewpoint generation algorithm leveraging the latest observation, the ROI, and the trained ROI-ScoreNet (line 7). Then, the robot moves the in-hand camera to the desired viewpoint $v^t$ using the RRT-Connect motion planner \cite{844730} and returns the captured image $I^t$, the new scene representation $S^t$ and updated ROI coverage score $c$ (line 8). This is done through the \texttt{observe($v$)} function, which computes the newly observed region through the depth information. Since there are view-blocking objects that occlude the view to the ROI, the system now moves on to the object manipulation phase. Based on the reverse order of the BScore associated with each observed object, the robot arm relocates them to a feasible region outside the ROI before making another observation at the previous viewpoint $v^{t-1}$ (lines 11 - 15). The previous viewpoint is used because it aligns better with our object selection strategy, as the blocking score is calculated based on it. The object manipulation phase ends when all detected objects are removed or the ROI is already fully observed (line 10). All objects removed from the scene will be placed back in their original region so that the scene stays intact. Finally, as long as there are still unobserved regions in the ROI, the pipeline enters the next round, starting from the next best viewpoint selection through the GMM-MPC method(line 7). After the ROI is fully explored, the user is provided the minimal number of images $I_{\{t\}}$, which reveals all the details in the request region so that their request is fulfilled (line 20).

\begin{algorithm}
    \caption{Language-guided Active Sensing Approach}
    \begin{algorithmic}[1]
        \State{$S^0_o\leftarrow \emptyset; c\leftarrow 0; t \leftarrow 1$} \Comment{initialization}
        \State{$v^0; f_\theta$} \Comment{initial viewpoint and ROI-ScoreNet function}
        \State{$I^0 \leftarrow \text{observe}(v^0)$} \Comment{Initial image for user to select ROI}
        \State{$prompt\leftarrow \text{UserInput}(I^o)$} \Comment{ROI description from user}
        \State{$S_{ROI}\leftarrow$ ROI$\_$generation$(prompt)$}
        \While{$t\leq T_{max}$ or $c\leq c_{max}$}
            \State{$v^t \leftarrow $ GMM-MPC$(S^{t - 1},S_{ROI},f_\theta)$}
            \State{$(I^t, S^t, c) \leftarrow \text{observe}(v^t)$} \Comment{new observation}
            \State{$(o,h_o)_{\{N^t\}}\leftarrow$ sorted(BScore($S^t, S_{ROI}, v^t$))}
            
                \While{$c\leq c_{max}$ \textbf{and} $N^t > 0$}
                \State{$t \leftarrow t + 1$}
                \State{$o^t \leftarrow \text{pop}((o,h_o)_{\{N_{t-1}\}})$} \Comment{view-blocking object}
                \State{\text{object}\_\text{removal}($o^t$)}
                 \State{$(I^t, S^t, c) \leftarrow \text{observe}(v^{t-1})$}
                \State{$(o,h_{o})_{\{N^t\}}\leftarrow$  sorted(BScore($S^t, S_{ROI}, v^{t}$))}
            \EndWhile
            \State{scene$\_$restoration()} \Comment{put removed objects back}
            \State{$t\leftarrow t+1$}
        \EndWhile
        \State{\Return{$S^{t}_o, I_{\{t\}}$}} \Comment{final observed scene and image sets}
    \end{algorithmic}
    \label{NG-AS algorithm}
\end{algorithm}
\vspace{-0.2in}

\begin{figure*}[t]
\includegraphics[width=\textwidth]{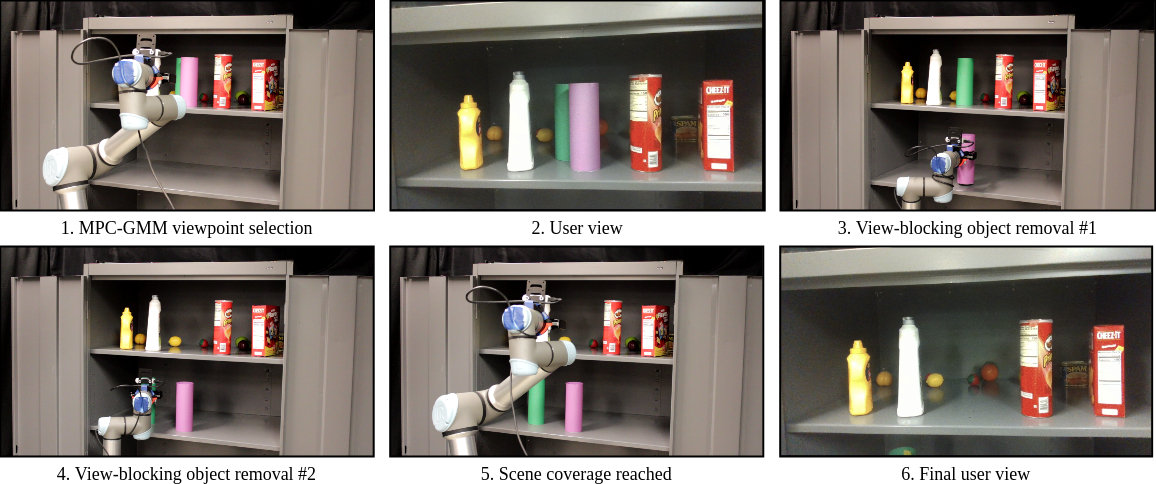}
\centering
\caption{This figure shows the real-world demonstration of our pipeline. The prompt given to the system is ``Show me behind the purple cylinder". From the top left figure, the robot starts by choosing the viewpoint that's roughly in front of the object of interest. The view-blocking object selection algorithm identified two objects to remove. By removing the two objects, the desired coverage is reached and the scene is restored.}
\label{fig:real}\vspace{-0.15in}\end{figure*}

\subsection{Implementation Details}
In this section, we provide the implementation details of our data collection procedure for the ROI ScoreNet training. First, we set up a virtual shelf environment in the IsaacGym simulator \cite{isaacgym}. The shelf has a randomized width of 70-140 cm, height of 25-40 cm, and depth of 50-100 cm. The position of the shelf relative to the robot is also randomized. The shelf can be 30-70 cm away from the robot and 30-60cm above ground level. Inside the shelf, 10-20 different objects are placed with randomized positions and orientations. The objects are further grouped into two categories: large and small. The large objects are placed towards the front for task complexity, i.e., to block a majority of the view for any camera angle. The region of interest $S_{ROI}$ is chosen by selecting a random anchor object in the scene and a random direction relative to said object from \{left, right, front, behind\}. From these randomly instantiated scenarios, we collect 10,000 different pairs of $(S^t, S_{ROI}, v^t, c^{t+1})$. These pairs are used for training our score function with 0.8, 0.1, and 0.1 data split for the training, validation, and testing set, respectively.


\begin{figure*}[t]
\includegraphics[width=\textwidth]{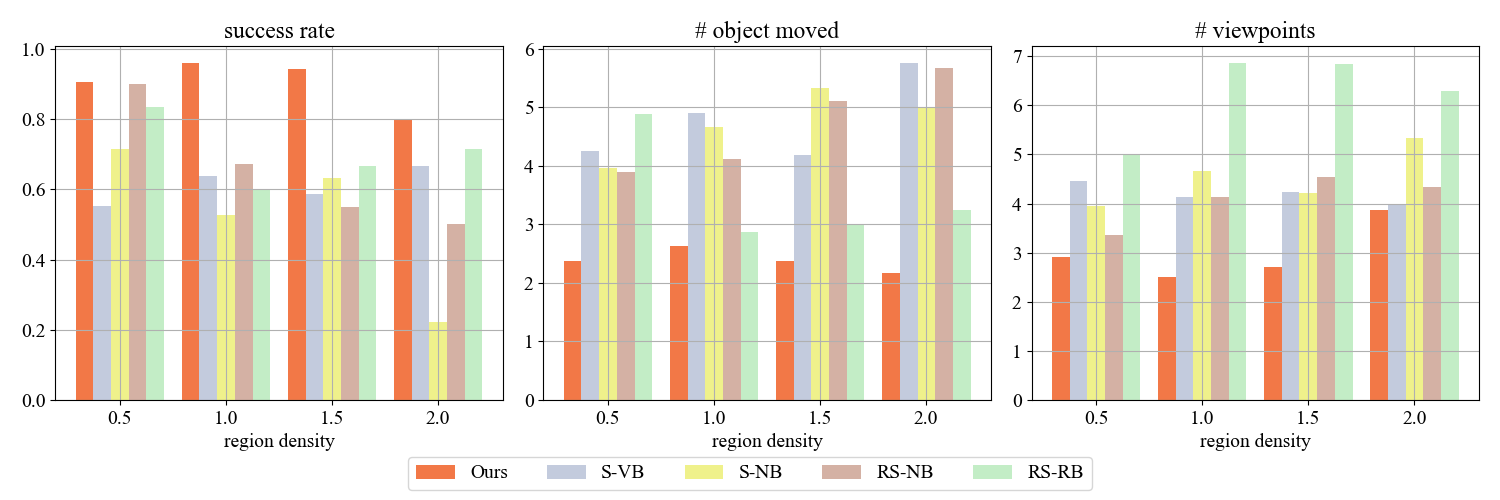}
\centering
\vspace{-0.2in}\caption{This figure shows the performance of our method against various baselines under different region densities. Our method outperforms other baselines in all densities in terms of success rate and efficiency.}
\label{graph}
\vspace{-0.2in}
\end{figure*}
\section{Results}
In this section, we present the result and analysis of the following experiments: 1) Comparison experiments to evaluate the performance of our proposed pipeline against multiple baselines in various randomized environments, which also reveals the importance of each component in our methodology; 2) Real-world experiments to demonstrate the sim2real transfer capabilities.
\subsection{Simulation Experiments}
The simulated experiments against the baselines are conducted in 100 randomly generated scenes. The parameters of the environment that are randomized include the following: the robot's distance from the scene, scene dimensions, the objects placed in the scene, and their poses. For each method, the ROI exploration ends either when the ROI is fully observed or the number of unique viewpoints exceeds a certain threshold, where the former is considered successful while the latter is a failure. The following metrics are collected and utilized to evaluate the methods :
\begin{itemize}
    \item Success Rate (SR): An experiment is considered successful if the ROI coverage rate exceeds the threshold using less than 8 unique viewpoints. 
    \item Number of objects moved (\# Object-moved): It tracks the number of objects that are removed from the scene before reaching the coverage threshold.
    \item Number of viewpoints (\# Viewpoints): It records the number of unique viewpoints the robot has moved to before reaching the coverage threshold.
    \item Time: It shows the time consumed before reaching the coverage requirement, including the planning time of the various components and the time for the robot to move to the different viewpoints.
\end{itemize}
The ROI coverage threshold $c_{max}$ is set to be 80\% based on empirical observations. With an 80\% coverage, the scene and the objects within it can be seen with sufficient clarity. The remaining 20\% is attributed to the lack of information inside large objects and the remaining regions 
near the back surface of the ROI where no other objects exists.\par
The following baseline algorithms are designed as comparisons to highlight the performance of various modules. Note that since our approach is the first method to solve language-guided active sensing via object manipulation problem, we utilize some heuristics like nearest object first or random related methods in the baselines. The baselines differ from our approach regarding the ROI coverage score prediction method and the object removal strategy. We use the following shorthand notation to describe the various aspects of the baselines: For the coverage rate prediction network, (RS) represents the proposed ROI ScoreNet while (S) is the original ScoreNet formulation from \cite{10101696}. On the other hand, for the object removal strategy, (VB) stands for the proposed blocking-score-based method, (RB) indicates a random object selection method, and (NB) greedily chooses the nearest object from the camera viewpoint. Hence, the four constructed baseline methods include ROI-ScoreNet with random object selection (RS-RB), ROI-ScoreNet with nearest object selection (RS-NB), ScoreNet with blocking-score-based object removal (S-VB), and ScoreNet with nearest object selection (S-NB). Furthermore, all methods mentioned above use our GMM-MPC for viewpoint generation.\par
Table \ref{tab:comparison} shows the performance of our approach against the baselines. In all, our method achieves the highest success rate and lowest number of object manipulations and viewpoints, outperforming the baselines in all metrics. The random object baseline (RS-RB) significantly underperformed, highlighting that random object removals from ROI do not yield the best observability of the scene. The nearest object selection method (RS-NB) failed to achieve a higher success rate, indicating that the nearest object to the in-hand camera is not always the view-blocking object. Furthermore, we observed that in the GMM-MPC with ROI-Scorenet, the viewpoints are generated relatively farther away from the ROI with appropriate angles, leading to better scene coverage. 
Hence, the objects closer to the camera are often not the view-blocking objects. Thus, removing them does not help to increase the ROI coverage rate. On the other hand, the original ScoreNet formulation in \cite{10101696} yields a low success rate since it considers the whole scene instead of just the ROI. Even with our proposed object selection algorithm, the coverage can hardly be improved because the coverage of the viewpoint generated did not overlap with the ROI. Lastly, when both viewpoint generation and object selection methods are replaced, i.e., (S-NB), the success rate and other metrics are further degraded as expected. \\
Figure \ref{graph} shows the visual comparison between our method and the baselines under varying region densities. Since the scene size, region of interest, and objects are randomized in each experiment trial, differentiating these factors can provide more insight into the performance of the various methods. Thus, we decided to stratify the results using region density, which is defined as the ratio of the number of objects in ROI and the ROI volume. As apparent from the graph, the difficulty of active ROI sensing increases with region density since more objects can block the view. It can also be seen from the graph that our method outperforms other baselines in terms of success rate, number of object manipulations, and number of viewpoints under all region densities.

\begin{table}
   \centering
   \scalebox{0.85}{
       \begin{tabular}{ccccc}\toprule
           \multirow{2}{*}{Algorithms}&\multicolumn{3}{c}{Performance Metrics}\\ \cmidrule{2-5}
           &\multicolumn{1}{c}{SR $(\%)$ $\uparrow$}&\multicolumn{1}{c}{\# Obj-moved $\downarrow$}&\multicolumn{1}{c}{\# Viewpoints $\downarrow$}&\multicolumn{1}{c}{Time (s) $\downarrow$}\\
           \midrule
           \multirow{1}{*}{Ours}& \multirow{1}{*}{$\bm{93}$} & \multirow{1}{*}{\bm{$2.69\pm 1.72$}}  &\multirow{1}{*}{\bm{$2.79\pm 2.42$}} &\multirow{1}{*}{\bm{$61.3\pm 44.4$}}\\
           \multirow{1}{*}{RS-RB}& \multirow{1}{*}{$66$} & \multirow{1}{*}{$3.62\pm 3.54$}  &\multirow{1}{*}{$6.60\pm3.67$} &\multirow{1}{*}{$265.11\pm 190.35$}\\
           \multirow{1}{*}{RS-NB}& \multirow{1}{*}{$65$} & \multirow{1}{*}{$4.78\pm 2.03$}  &\multirow{1}{*}{$4.16\pm 1.46$} &\multirow{1}{*}{$145.54\pm 62.93$}\\
           \multirow{1}{*}{S-VB}& \multirow{1}{*}{$60$} & \multirow{1}{*}{$4.68\pm 1.61$}  &\multirow{1}{*}{$4.25\pm 1.46$} &\multirow{1}{*}{$152.59\pm 58.76$}\\ 
           \multirow{1}{*}{S-NB}& \multirow{1}{*}{$58$} & \multirow{1}{*}{$4.84\pm 1.83$}  &\multirow{1}{*}{$4.39\pm 1.48$} &\multirow{1}{*}{$159.40\pm 61.89$}\\ 
           \bottomrule
    \end{tabular}
    }
    \caption{This table shows the results of the experiments. Our proposed method can be seen to have the highest success rate and is most efficient in object removal and viewpoint generation. Due to this efficiency, the processing time of our method is significantly lower than the other baselines. } \label{tab:comparison}
        \vspace{-0.2in}
\end{table}

\subsection{Real Robot Experiments}
We created three scenes with different user prompts to test the effectiveness of our pipeline in real-world settings. In real settings, we utilize an image segmentation model to generate the object point clouds. The image segmentation model we use is grounding-SAM \cite{IDEA-Research}. This model utilizes the segment-anything \cite{kirillov2023segment} model and grounding-DINO \cite{liu2023grounding} to produce segmentation masks from language inputs.\\
The scene settings and the prompts in the real experiments are crafted with the intent of testing the different capabilities of the proposed method. The prompt in the first scene, as can be seen in Fig. \ref{fig1:pipeline}, is ``Show me to the left of the Pringles can." Our pipeline relocates two objects to reveal the majority of the requested area. In the second experiment shown in Fig. \ref{fig:real}, the robot is asked to explore the region behind the purple cylinder. This scene is significant in showing the effectiveness of our pipeline since the green cylinder is only partially visible. It can only be fully seen and identified as a view-blocking object after removing the purple one. The complete execution processes of all real experiments are available in the supplementary videos. In all, these successful real-world experiments demonstrate the robustness of our system under adverse lighting and can generalize to environments with daily-life objects with arbitrary placements.

\section{Conclusion}
This paper presents a novel method for language-guided active sensing via object manipulation in unknown, cluttered environments. Our approach first identifies an ROI given by a user via natural language. Then, it explores the ROI through a unique neural network-based viewpoint generation algorithm and a specially designed view-blocking object manipulation strategy. The efficacy of our method is first tested in the simulation environments and further verified in real-world scenarios. Comparative analyses indicate that our strategy surpasses other benchmark methods regarding the success rate and the number of object interactions. The real-world experiments justify the sim2real transfer capability of our pipeline, making it suitable for direct deployment. In our future work, we plan to build an end-to-end deep learning approach that optimizes viewpoints and object relocations to maximize the ROI coverage. We believe such an approach can realize even faster execution for real-time applications.







\bibliographystyle{IEEEtran}
\bibliography{root}

\end{document}